\documentclass[aps,12pt,epsf]{revtex4}

\usepackage{amsmath}
\usepackage{amssymb}
\usepackage{graphicx}
\usepackage{comment}
\usepackage{pseudocode}

\begin{document}

\title{Learning image transformations without training examples.}

\author{Sergey \surname{Pankov}}
\affiliation{Harik Shazeer Labs, Palo Alto, CA 94301}

\begin{abstract}

The use of image transformations is essential for efficient modeling and learning of visual data. But the class of relevant transformations is large: affine transformations, projective transformations, elastic deformations, ... the list goes on. Therefore, learning these transformations, rather than hand coding them, is of great conceptual interest. To the best of our knowledge, all the related work so far has been concerned with either supervised or weakly supervised learning (from correlated sequences, video streams, or image-transform pairs). In this paper, on the contrary, we present a simple method for learning affine and elastic transformations when no examples of these transformations are explicitly given, and no prior knowledge of space (such as ordering of pixels) is included either. The system has only access to a moderately large database of natural images arranged in no particular order.

\end{abstract}

\maketitle

\section{Introduction}
\label{intro}

Biological vision remains largely unmatched by artificial visual systems across a wide range of tasks. Among its most remarkable capabilities are the aptitude for unsupervised learning and efficient use of spatial transformations. Indeed, the brain's proficiency in various visual tasks seems to indicate that some complex internal representations are utilized to model visual data. Even though the nature of those representations is far from understood, it is often presumed that learning them in an unsupervised manner is central to the biological neural processing \cite{barlow89} or, at very least, highly relevant for modeling neural processing computationally \cite{unsupervisedlearning99, zemel93, oja02}. Likewise, it is poorly understood how the brain implements various transformations in its processing. Yet it must be clear that the level of learning efficiency demonstrated by humans and other biological systems can only be achieved by means of transformation-invariant learning. This follows, for example, from an observation that people can learn to recognize objects fairly well from only a small number of views.

Covering both topics (unsupervised learning and image transformations) at once, by way of learning transformations without supervision, appears interesting to us for two reasons. Firstly, it can potentially further our understanding of unsupervised learning: what can be learned, how it can be learned, what are its strengths and limitations. Secondly, the class of transformations important for representing visual data may be too large for manual construction. In addition to transformations describable by a few parameters, such as affine, the transformations requiring infinitely many parameters, such as elastic, are deemed to be important \cite{belongie02}. Transformations need not be limited to spacial coordinates, they can involve temporal dimension or color space. Transformations can be discontinuous, can be composed of simpler transformations, or can be non-invertible. All these cases are likely to be required for efficient representation of, say, an animal or person. Unsupervised learning opens the possibility of capturing such diversity.

A number of works have been devoted to learning image transformations \cite{rao97, rao99, miao07, memisevic07, memisevic10, sohldickstein10}. Other works were aimed at learning perceptual invariance with respect to the transformations \cite{foldiak91, wallis93, stringer06}, but without explicitly extracting them. Often, no knowledge of space structure was assumed (such methods are invariant with respect to random pixel permutations), and in some cases the learning was termed unsupervised. In this paper we adopt a more stringent notion of unsupervised learning, by requiring that no ordering of an image dataset be provided. In contrast, the authors of the cited references  considered some sort of temporal ordering: either sequential (synthetic sequences or video streams) or pairwise (grouping original and transformed images). Obviously, a learning algorithm can greatly benefit from temporal ordering; just like ordering of pixels opens the problem to a host of otherwise unsuitable strategies. Ordering of images provides explicit examples of transformations. Without ordering, no explicit examples are given. It is in this sense that we talk about learning without (explicit) examples. 

The main goal of this paper is to demonstrate learning of affine and elastic transformations from a set of naturals images by a rather simple procedure. Inference is done on a moderately large set of random images, and not just on a small set of strongly correlated images. The latter case is a (simpler) special case of our more general problem setting. 

The possibility of inferring even simple transformations from an unordered dataset of images seems intriguing in itself. Yet, we think that dispensing with temporal order has a wider significance. Temporal proximity of visual percepts can be very helpful for learning some transformations but not others. Even the case of 3D rotations will likely require generation of hidden parameters encoding higher level information, such as shape and orientation. That will likely require processing a large number of images off-line, in a batch mode, incompatible with temporal proximity.

The paper is organized as follows. A brief overview of related approaches is given in section \ref{related}. Our method is introduced in section \ref{learning}. In section \ref{results} the method is tested on a synthetic and natural sets of random images. In section \ref{discussion} we conclude with discussion of limitations and possible extensions of the current approach, outlining a potential application to the learning of 3D transformations.

\section{Related work}
\label{related}

It is recognized that transformation invariant learning, and hence transformations themselves, possess great potential for artificial cognition. Numerous systems, attempting to realize this potential, have been proposed over the last few decades. In most cases the transformation invariant capabilities were bult-in. In the context of neural networks, for example, translational invariance can be built-in by constraining weights of connections \cite{fukushima80, lecun90}. Some researchers used natural image statistics to infer the underlying structure of space without inferring transformations. For example, ideas of redundancy reduction applied to natural images, such as independent component analysis or sparse features, lead to unsupervised learning of localized retinal receptive fields \cite{atick93} and localized oriented features, both in spatial \cite{olshausen96} and spatio-temporal \cite{vanhateren98} domains.

As we said, transformation (or transformation-invariant) learning has so far been implemented by taking advantage of temporal correlation in images. In Refs. \cite{foldiak91, wallis93, stringer06} transformation-invariant learning was achieved by incorporating delayed response to stimuli into Hebbian-like learning rules.

By explicitly parametrizing affine transformations with continuous variables it was possible to learn them first to linear order in Taylor expansion \cite{rao97} and then non-perturbatively as a Lie group representation \cite{rao99, miao07, sohldickstein10}. In the context of energy-based models, such as Boltzmann machines, transformations can be implemented by means of three-way interactions between stochastic units. The transformations are inferred by learning interaction strengths \cite{memisevic07, memisevic10}. In all these cases the corresponding algorithms are fed with training examples (of possibly several unlabeled types) of transformations. Typically, images do not exceed 40 $\times$ 40 pixels in size.

Below we demonstrate that image transformations can be learned without supervision, and without temporal ordering of training images. We consider both synthetic and natural binary images, achieving slightly better result for the synthetic set. Transformations are modeled as pixel permutations in 64 $\times$ 64 images. We see many possible modifications to our algorithm enabling more flexible transformation representation, more efficient learning, larger image sizes, etc. These ideas are left for future exploration. In the current manuscript, our main focus is on showing the feasibility of the proposed strategy in its basic incarnation.

\section{Learning transformations from unordered images}
\label{learning}

The basic idea behind our algorithm is extremely simple. Consider a pair of images and a transformation function. Introduce an objective function characterizing how well the transformation describes the pair, treating it as an image-transform pair. Minimize the value of the objective function across a subset of pairs by modifying the subset and the transformation incrementally and iteratively. The subset is modified by finding better-matching pairs in the original set of images, using fast approximate search. We found that a simple hill climbing technique was sufficient for learning transformations in relatively large 64 $\times$ 64 images. Bellow we describe the algorithm in more detail.

\subsection{Close match search}
\label{closematch}

Let $\mathcal S$ be a set of binary images of size $L\times L$. We sometimes refer to $\mathcal S$ as the set of random images. The images are random in the sense that they are drawn at random from a much larger set $\mathcal N$, embodying some aspects of natural image statistics. For example, $\mathcal N$ could be composed of: a) images of a white triangle on black background with integer-valued vertex coordinates ($|\mathcal N|=L^3/3!$ images), b) $L\times L$ patches of (binarized) images from the Caltech-256 dataset \cite{griffin07}. We will consider both cases. Notice that our definition of $\mathcal S$ implies that it needs to be sufficiently large to contain pairs of images connectable by a transformation of interest. Otherwise such transformation cannot be learned.

To learn a transformation at $L=64$ we will need $|\mathcal S|$ to be in the order of $10^4-10^5$, with the number of close match searches in the order of $10^5-10^6$. Clearly, it is crucial to employ some efficient search technique. In a wide class of problems a significant speedup can be achieved by abandoning exact nearest neighbor search in favor of approximate nearest neighbor search, with little loss in quality of performance. Our problem appears to belong to this class. Therefore, approximate algorithms, such as best bin first \cite{beis97} or locality sensitive hashing (LSH) \cite{gionis99}, are potential methods of choice. LSH seems especially suitable thanks to its ability to deal with high-dimensional data, like vectors of raw pixels. On the flip side, LSH requires estimation of optimal parameters, which is typically done with an implicit assumption that the query point is drawn from the same distribution as the data points. Not only is that not the case here, the query distribution itself changes in the course of the algorithm run. Indeed, in our case the query is the image transform under the current estimate of the transformation. It gradually evolves from a random permutation, to something approximating a continuous 2D transformation. To avoid these complications we opt for storing images in binary search trees, while also creating multiple replicas of the tree to enhance performance in the spirit of LSH. Details are given below, but first we introduce a few notations.

Let a $L\times L$ binary image be represented by a binary string ${\bf x}\equiv x_1...x_{L^2}$, where $x_i$ encodes the color (0=black, 1=white) of the pixel in the $i$-th position (under some reference ordering of pixels). Let $o$ be an ordering of pixels defined as a permutation relative to the reference ordering. Given $o$, the image is represented by the string ${\bf x}(o)\equiv x_{o(1)}...x_{o(L^2)}$. We will refer to an image and its string representation interchangeably, writing ${\bf x}_I(o)$ to denote an image $I$.

Let $B(o)$ be a binary search tree that stores images $I\in\mathcal S$ according to (lexicographic) order of ${\bf x}_I(o)$. Rather than storing one image per leaf, we allow a leaf to contain up to $m$ images (that is any subtree containing up to $m$ images is replaced by a leaf). We construct $l$ versions of the tree data structure, each replica with a random choice of $o_i$, $i=1,...,l$. This replication is intended to reduce the possibility of a good match being missed. A miss may happen if a mismatching symbol (between a query string and a stored string) occurs too soon when the tree is searched. Alternatively, one could use a version of A* search, as in the best bin first algorithm, tolerating mismatched symbols. However, our empirical results suggest that the approach benefits from tree replication, possibly because information arrives from a larger number of pixels in this case.

To find a close match to an image $I$, we search every binary tree $B(o_i)$ in the usual way, using ${\bf x}_I(o_i)$ as query string. The search stops at a node $n$ if: a) $n$ is a leaf-node, or b) search cannot proceed further ($n$ lacks an appropriate branch). All the images from the subtree rooted at $n$ are returned. In the final step we compute distance to every returned candidate and select the closest to $I$ image. 

In short, in our close match search algorithm we use multiple binary search trees, with distinct trees storing images in distinct random orderings of pixels. The described approximate search yields a speedup of $|{\mathcal S}|/ml$ over the exact nearest neighbor search. For the values $m=5$ and $l=10$ that we used in our experiments (see section \ref{results}), the speedup was about $10^2-10^3$.

\subsection{Transformation optimization}

We define image transformation $T$ as a permutation of pixels $t$. That is $T{\bf x}={\bf x}(t)$. Despite obvious limitations of this representation for describing geometric transformations, we will demonstrate its capacity for capturing the essence of affine and elastic transformations. To be precise, our method in the current formulation can only capture a volume-preserving subset of these transformations. But removing this limitation should not be too difficult (see section \ref{discussion} for some discussion).

We denote a pair of images as $(I,I')$ or $({\bf x}_I,{\bf x}_{I'})$. The Hamming distance between strings $\bf x$ and $\bf x'$ is defined as $d({\bf x},{\bf x}')\equiv\sum_i(x_i-x'_i)^2$. The objective function $d_T$, describing how well a pair of images is connected by the transformation $T$, is defined as:\begin{equation}
d_T(I,I')\equiv d(T{\bf x}_I,{\bf x}_{I'}).
\label{dT}
\end{equation}
Thus, the objective function uses the Hamming distance to measure dissimilarity between the second image and the transform of the first image.

We will be minimizing $d_T$ across a set of pairs, which we call the pair set and denote it $\mathcal{P}$. The objective function $D_T$ over $\mathcal P$ is defined as:
\begin{equation}
D_T\equiv \sum_{p\in\mathcal{P}}d_T(p),
\label{DT}
\end{equation}
where we used a shorthand notation $p$ for a pair from the pair set. 
We refer to the minimization of $D_T$ while $\mathcal{P}$ is fixed and $T$ changes as the transformation optimization. We refer to the minimization of $D_T$ while $T$ is fixed and $\mathcal{P}$ changes as the pair set optimization.

In the transformation optimization phase the algorithm attempts to minimize $D_T$ by incrementally modifying $T$. A simple hill climbing is employed: a pair of elements from $t$ are exchanged at random, the modification is accepted if $D_T$ does not increase. A transformation modification effects every transformed string ${\bf x}(t)$ from $\mathcal P$. However, there is an economical way of storing the pair set that makes computation of $D_T$ particularly fast. Consider the first images of all pairs from $\mathcal P$. Consider a matrix whose rows are these images. Let ${\bf x}_i$ be $i$-th column in this matrix. Define similarly ${\bf x}'_i$, by arranging the second images in the same order as their pair counterparts. The objective function expressed through these vector notations then reads:
\begin{equation}
D_T=\sum_{i=1}^{L^2}({\bf x}_{t(i)}-{\bf x}'_i)^{\top}({\bf x}_{t(i)}-{\bf x}'_i).
\label{DT1}
\end{equation}
If the $i$-th and $j$-th elements of $t$ are exchanged, then the corresponding change $(\Delta D_T)_{ij}$ in the objective function reads:
\begin{equation}
(\Delta D_T)_{ij}=2({\bf x}_{t(i)}-{\bf x}_{t(j)})^{\top}({\bf x}'_i-{\bf x}'_j),
\label{deltaDT}
\end{equation}
which involves computing four terms of the form ${\bf x}_{a}^{\top}{\bf x}'_b$. For a binary image, as in our case, the vectors are binary strings and their dot-products can be computed efficiently using bitwise operations. Notice also that the vectors in Eq.(\ref{DT1}) are unchanged throughout the transformation optimization phase, only $t$ is updated. A transformation optimization phase followed by a pair set optimization phase constitutes one iteration of the algorithm. There are $n_t$ attempted transformation modifications per one iteration.

\subsection{Pair set optimization}
\label{pairoptimization}

The goal of the pair set optimization is twofold. On one hand, we want $\mathcal P $ to contain pairs that minimize $D_T$. On the other hand, we would like to reduce the possibility of getting stuck at a local minimum of $D_T$. To achieve the first goal, we update $\mathcal P$ by adding new pairs, ranking all pairs according to $d_T$ and removing the pairs with highest $d_T$. To add a new pair $(I,I')$ we pick image $I$ at random from $\mathcal S$, then search for $I'$ as a close match to $T{\bf x}_I$. To achieve the second goal, we add stochastic noise to the process by throwing out random pairs from the pair set.

We denote $n_n$ and $n_r$ the number of newly added pairs and the number of randomly dropped pairs respectively, both per one iteration. After $n_n$ pairs are added and $n_r$ pairs are dropped, we remove $n_n-n_r$ pairs with highest $d_T$, so that the number of pairs $|\mathcal P|$ in the pair set remains unchanged.

\subsection{Summary of the algorithm}

We briefly summarize our algorithm in Alg. \ref{minimized}. First, $T$ and $\mathcal P$ are randomly initialized, then the procedure \textsc{minimizeD}($T,\mathcal P$) is called. It stochastically minimizes $D_T$ by alternating between transformation optimization and pair set optimization for total of $n_i$ iterations.

\begin{pseudocode}[ruled]{minimizeD}{T,\mathcal P}
\FOR 1\TO n_i
\DO\BEGIN
	\FOR 1 \TO n_t
	\DO\BEGIN
		(i,j)\GETS (\CALL{rand}{L^2},\CALL{rand}{L^2})\\
		\CALL{exchange}{i,j,T}\\
		\IF \CALL{deltaD}{i,j,T,{\mathcal P}} > 0
		\THEN \CALL{exchange}{i,j,T}\\
	\END\\
	\CALL{addPairs}{n_n,{\mathcal P}}\\
	\CALL{dropPairs}{n_r,{\mathcal P}}\\
	\CALL{removePairs}{n_n-n_r,{\mathcal P}}\\
\END\\
\label{minimized}
\end{pseudocode}

Calls to other procedures should be self-explanatory in the context of the already provided description: \textsc{rand}$(n)$ generates a random integer in the interval $[1,n]$, \textsc{exchange}$(i,j,T)$ exchanges the $i$-th and $j$-th elements of $t$, \textsc{deltaD}$(i,j,T,{\mathcal P})$ computes $(\Delta D_T)_{ij}$ according to Eq.(\ref{deltaDT}); finally, \textsc{addPairs}($n,{\mathcal P}$), \textsc{dropPairs}($n,{\mathcal P}$) and \textsc{removePairs}($n,{\mathcal P}$) adds random, drops random, and removes worst performing (highest $d_T$) $n$ pairs respectively, as explained in subsection \ref{pairoptimization}.

As is often the case with greedy algorithms, we cannot provide guarantees that our algorithm will not get stuck in a poor local minimum. In fact, due to the stochasticity of the pair set optimization, discussing convergence itself is problematic. Instead, we provide convincing empirical evidence of the algorithm's efficacy by demonstrating in the next section how it correctly learns a diverse set of transformations.

\section{Results}
\label{results}

We tested our approach on two image sets: a) synthetic set of triangles, b) set of natural image patches. These experiments are described below.

\subsection{Triangles}

Edges and corners are among the commonest features of natural scene images. Therefore a set of random triangles is a good starting point for testing our approach. 
The set $\mathcal S$ is drawn from the set $\mathcal N$ of all possible white triangles on black background, whose vertex coordinates are restricted to integer values in the range $[0,L)$. For convenience, we additionally restricted $S$ to contain only images with at least 10\% of minority pixels. This was done to have better-balanced search trees, and also to increase informational content of $\mathcal S$, since little can be learned from little-varying images.

Our goal was to merely demonstrate that this approach can work, therefore we did not strive to find best possible parameters of the algorithm. Some parameters were estimated\footnote
{A rough estimate of set sizes goes as follows. Say, we want to infer a transformation at a resolution of $\varepsilon$ pixels. A random triangle will have a match in $\mathcal S$ within this resolution if $|\mathcal S| \ge (L/\varepsilon)^3$. The transformation will be represented by $\mathcal P$ down to the required resolution if $|\mathcal P| \ge L/\varepsilon$. Even more hand-waving estimate of the algorithm loop sizes goes as follows. To ensure the incremental character of changes we need: $n_t\ll L^4$, $n_n\ll |\mathcal P|$. To counter the threat of poor local minima we need $\mathcal P$ to be renewed many times, but not too fast, so $n_i \gg |\mathcal P|/n_r$ and $n_r \ll n_n$. These estimates should be viewed as no more than educated guesses.}, 
and some were found by a bit of trial and error. The parameters we used were: $L=64$, $m=5$, $l=10$, $|\mathcal S|=30000$, $|\mathcal P|=200$, $n_t=10000$, $n_n=10$, $n_r=1$, $n_i=3000$.

We want to show that the algorithm can learn without supervision multiple distinct transformations that are representative of $\mathcal S$. The simplest strategy is to generate transformations starting from random $T$ and $\mathcal P$, eliminating samples with higher $D_T$ to minimize the chance of including solutions from poor local minima. For more efficiency, compositions of already learned transformations can be used as initial approximations to $T$. Compositions can also be chosen to be far from learned samples. We found that for $L=64$ poor solutions occur rarely, in less than approximately 10\% of cases. By poor we mostly mean a transformation that appears to have a singularity in its Jacobian matrix. We chose to generate about three quarters of transformations from nonrandom initial $T$, setting $n_i=1000$ in such cases. Half of all generated samples were kept. In this way the algorithm learned about half a hundred transformations completely without supervision.

\begin{figure}
\includegraphics[scale=.125]{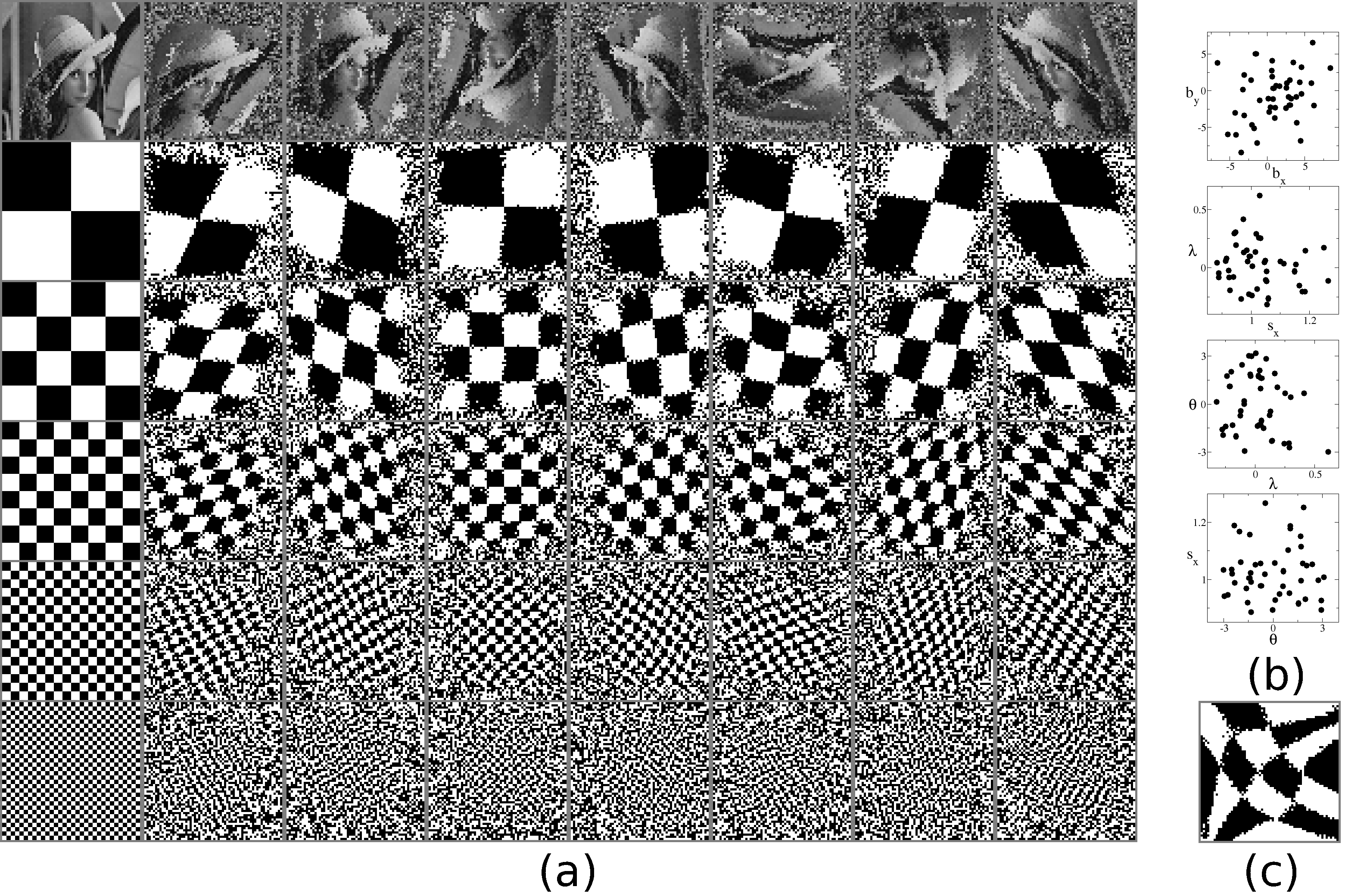}
\caption{Visualization of learned transformations for the set of triangles (a,b) and natural image patches (c). a) Selected examples of affine transformations. b) Set of learned transformations projected onto various planes in parameter space, top to bottom: $b_x$-$b_y$, $s_x$-$\lambda$, $\lambda$-$\theta$, $\theta$-$s_x$. c) A typical transformation visualized by $4\times4$ checkerboard pattern.}
\label{transforms}
\end{figure}

All learned transformations looked approximately affine. Selected representative examples (for better quality additionally iterated with $n_i=5000$, $|\mathcal P|=300$ and $|\mathcal S|=100000$) are shown in Fig.(\ref{transforms}.a). Since the human eye is very good at detecting straight parallel lines, we deemed it sufficient to judge quality of the learned affine transformations by visual inspection of the transforms of appropriate patterns. The transformations are visualized by applying them to a $L\times L$ portrait picture and checkerboard patterns with check sizes 32, 16, 8, 4 and 2. Since the finest checkerboard pattern is clearly discernible, we conclude that the achieved resolution is no worse than 2 pixels. With our choice of representing transformations as pixel permutations it is difficult to expect a much better resolution. Other consequences of this choice are: a) all the captured transformations are volume preserving, b) there are white-noise areas that correspond to pixels that should be mapped from outside of the image in a proper affine transformation. Nonetheless, this representation does capture most of the aspects of affine transformations. To better illustrate this point we plot in Fig(\ref{transforms}.b) values of various parameters of all the learned examples. The parameters of an affine transformation $\xi'=A\xi+b$ are computed using:
\begin{equation}
A=C_{\xi'\xi} (C_{\xi\xi})^{-1},\qquad b=\mu_{\xi'}-A\mu_{\xi},
\label{affineparams}
\end{equation}
Where $C$ and $\mu$ are the covariance matrix and the mean: $C_{ab}=\langle ab^{\top}\rangle-\mu_a\mu_b^{\top}$ and $\mu_{a}=\langle a\rangle$. The averaging $\langle ... \rangle$ is weighted by a Gaussian with standard deviation $\sigma=.1L$ centered at $(L/2, L/2)$. The weighting is needed because our representation cannot capture an affine transformation far from the image center. We further parametrize $A$ in terms of a consecutively applied scaling $S$, transvection $\Lambda$ and rotation $R$. That is $A=R\Lambda S$ where:
\begin{equation}
S=\left( \begin{array}{cc}
s_x & 0 \\
0 & s_y
\end{array} \right),\qquad
\Lambda=\left( \begin{array}{cc}
1 & \lambda \\
0 & 1
\end{array} \right),\qquad
R=\left( \begin{array}{cc}
\cos{\theta} & -\sin{\theta} \\
\sin{\theta} & \cos{\theta}
\end{array} \right).
\end{equation}
The parameters $s_x$, $s_y$, $\lambda$ and $\theta$ expressed in terms of $A$ are: $s_x=\sqrt{A_{11}^2+A_{21}^2}$, $s_y={\rm Det(A)}/s_x$, $\lambda=(A_{11}A_{12}+A_{21}A_{22})/(s_x s_y)$ and $\theta={\rm atan2}(A_{21},A_{11})$, where ${\rm Det(A)}=A_{11}A_{22}-A_{21}A_{12}$. From Fig.(\ref{transforms}.b) we see that the parameter values are evenly distributed over certain ranges without obvious correlations. Unexplored regions of the parameter space correspond to excessive image distortions, with not many images in $\mathcal S$ connectable by such transformations at reasonable cost. Also, $|{\rm Det}(A)|$ across all transformations was found to be $.998 \pm .004$, validating our claim of volume preservation.

\subsection{Natural image patches}

In the second experiment we learned transformations from a set of natural images, derived from the Caltech-256 dataset \cite{griffin07}. The original dataset was converted to binary images using k-means clustering with $k=2$. Non-overlaping $L\times L$ patches with minority pixel fraction of at least 10\% were included in $\mathcal N$. We had $|\mathcal N|\approx 500000$. We used the following algorithm parameters: $L=64$, $m=5$, $l=10$, $|\mathcal S|=200000$, $|\mathcal P|=1000$, $n_t=10000$, $n_n=20$, $n_r=1$, $n_i=5000$.

Natural images are somewhat richer than the triangle set, consequently the transformation we learned were also richer. Typical transformation looked like a general elastic deformation, often noticeably differing from an affine transformation. White noise areas were much smaller or absent, while the resolution was lower at about 3 pixels. A typical example is shown in Fig.(\ref{transforms}.c).

\section{Discussion and conclusion}
\label{discussion}

In this paper we have demonstrated conceptual feasibility of learning image transformations from scratch: without image set or pixel set ordering. To the best of our knowledge learning transformations from unordered image dataset has never been considered before. Our algorithm, when applied to natural images, learns general elastic transformations, of which affine transformations are a special case.

For the sake of simplicity we chose to represent transformations as pixel permutations. This choice  restricted transformations by enforcing volume conservation. In addition, it adversely affected the resolution of transformations. We also limited images to binary form, although the learned transformations can be applied to any images. Importantly, we do not see any reason why our main idea would not be applicable in the case of a general linear transformation acting on continuously-valued pixels. In fact, the softness of continuous representation may possibly improve convergence properties of the algorithm. We plan to explore this extension, expecting it to capture arbitrary scaling transformations and to increase the resolution of learned transformations. 

Images that we considered were relatively large by standards of the field. For even larger images chances of getting trapped in a poor local minimum increase. To face this challenge we can propose a simple modification. Images should be represented by a random subset of pixels. Learning should be easy with a small initial size of the subset. In this way one learns a transformation at a coarse grained level. Pixels then are gradually added to the subset, increasing the transformation resolution, until all pixels are included. Judging from our experience, this modification will allow tackling much larger images.

It seems advantageous for the efficiency of neural processing to factor high dimensional transformations, such as affine transformations, into more basic transformations. How the learned random transformations can be used to that end is another interesting problem.

In our view, 3D rotations $R\eta \to \eta'$ can be learned in a similar fashion as we learned affine transformations, with orientations $\eta$ playing role of pixels in the current work. The problem however is much harder since we do not have direct access to hidden variables $\eta$. Indirect access is provided through projected transformations $A(R,\eta)$, where set of $A$ is presumed to have been learned (apart from its dependence on the arguments $R$ and $\eta$). We believe that the presence of multiple orientations in a given image and multiple images should constrain $R$ and $A$ sufficiently for them to be learnable. 

To conclude, we consider the presented idea of unsupervised learning of image transformation novel and valuable, opening new opportunities in learning complex transformations, possibly tackling such difficult cases as projections of 3D rotations.

\section{Acknowledgments}

We gratefully acknowledge many useful discussions with Noam Shazeer and Georges Harik.

\bibliographystyle{plain}

\bibliography{myrefs}

\end{document}